# IVGAE-TAMA-BO: A novel temporal dynamic variational graph model for link prediction in global food trade networks with momentum structural memory and Bayesian optimization

**Authors**: Sicheng Wang[1], Shuhao Chen[1], Jingran Zhou[1], Chengyi Tu[1*]

[1]School of Economics and Management, Zhejiang Sci-Tech University, Hangzhou, 310018, China.

*Corresponding author. Email: chengyitu1986@gmail.com

## Abstract

Global food trade is vital to food security and supply chain stability, yet its network structure evolves dynamically under geopolitical, economic, and environmental influences. These changes complicate the task of modeling and predicting future trade links. Thus, effectively capturing temporal patterns in food trade networks is essential for enhancing the accuracy and robustness of link prediction. This study introduces IVGAE-TAMA-BO, a novel dynamic graph neural network designed to model dynamic trade structures and predict future links in global food trade networks. To the best of our knowledge, this is the first work to introduce dynamic graph neural networks to this work, effectively improving predictive accuracy. Built upon the original IVGAE framework, the proposed model introduces a Trade-Aware Momentum Aggregator (TAMA) to capture the temporal evolution of food trade networks, jointly modeling short-term fluctuations and long-term structural dependencies. It incorporates a momentum-based structural memory mechanism, which enhances predictive stability and overall performance. Additionally, Bayesian optimization is employed to automatically tune key hyperparameters, improving the model's generalization across diverse trade scenarios. Extensive experiments on five crop-specific datasets demonstrate that IVGAE-TAMA substantially improves predictive performance over the static IVGAE and other dynamic baselines by effectively modeling temporal dependencies, while Bayesian optimization further enhances results in IVGAE-TAMA-BO through automatic hyperparameter tuning. These results suggest that the proposed framework offers a robust and scalable solution for structural prediction in global trade networks, with strong potential for applications in food security monitoring and policy decision support.

## 1 Introduction

Global food trade serves as a critical conduit linking production and consumption across regions, playing a vital role in ensuring food security, alleviating regional hunger, stabilizing agricultural markets, and optimizing resource allocation[1-4]. With the deepening of globalization and the growing interdependence of national agricultural systems, the international food trade network has become a foundational infrastructure underpinning the resilience of the global food system[5-7]. This network not only facilitates the transboundary flow of food commodities but also acts as a buffer against external shocks such as natural disasters, policy interventions, and geopolitical conflicts[8-10]. In the context of intensifying climate change and rising geopolitical uncertainty, effectively predicting and understanding the structural evolution of the global food trade network is of strategic importance for achieving the United Nations Sustainable Development Goals and safeguarding global food stability[11,12].

In traditional studies of international food trade, the Gravity Model[13] has been widely adopted in research for analyzing global food trade networks[14-16]. By incorporating macroeconomic variables such as national economic size and geographic distance, this model provides an intuitive and interpretable framework for modeling trade relationships[13]. However, with the advancement of complex network science, researchers have increasingly recognized that the structural and systemic properties of international trade networks cannot be fully captured by conventional econometric approaches. As a result, artificial intelligence techniques have gradually been introduced into trade network modeling, owing to their strengths in capturing complex relationships and uncovering high-order patterns. In recent years, Graph Neural Networks (GNNs) have achieved remarkable success in link prediction tasks, with widespread applications across domains such as social networks, knowledge graphs, and biological systems[17-20]. Leveraging their powerful ability to model high-order dependencies and intricate graph structures, GNNs have emerged as a leading approach in graph-based representation learning. Building on these advantages, GNNs have begun to be adopted for modeling international trade networks, aiming to capture the relational and structural characteristics of trade interactions between countries[21-24].

The global food trade network exhibits significant differences from other commodity trade networks in terms of structural characteristics and dynamic evolution. On one hand, as a strategic resource, food trade relationships are often influenced by external factors such as climate change, geopolitical conflicts, pandemics, and national intervention policies, resulting in high temporal instability in the network structure[3,25-27]. On the other hand, food import and export relationships for many countries are concentrated among a limited number of partners, leading to high structural sparsity and imbalance within the network[28-30]. These distinctive features impose greater demands on a model's capability for structural representation, temporal dependency modeling, and robustness. In the domain of food trade link prediction, Li et al. proposed the IVGAE-GA model[31] to address the issue of node over-smoothing and limited expressiveness caused by the highly sparse structure of food trade networks. By integrating a dynamic attention mechanism with a graph convolutional encoder, the model effectively captures both local and non-local structural features, enhancing the discriminative power and informativeness of node embeddings. Furthermore, the model incorporates a genetic algorithm for hyperparameter optimization, thereby improving its robustness and predictive accuracy in static sparse graphs.

Despite the initial success of graph neural networks (GNNs) in modeling food trade networks, several limitations remain. First, existing approaches are primarily based on static representations of food trade networks, overlooking the temporal evolution of trade relationships in practice. This static assumption hinders the model's ability to capture cross-year structural shifts and trend-related features. Second, international food trade networks typically exhibit strong structural inertia and long-term dependencies at the macro level. For example, trade relationships between major exporters and stable importers often persist over many years, forming path-dependent structural patterns. Therefore, capturing and retaining historical structural information is essential for enhancing model stability and predictive accuracy. However, most current GNN-based models lack an effective structural memory mechanism, making it difficult to reliably extract long-term relational patterns. As a result, they are vulnerable to short-term noise and local perturbations, limiting their ability to model the structural evolution of trade networks. Lastly, while the choice of hyperparameters plays a critical role in determining the overall performance of prediction models[32,33], existing studies often rely on manually tuned hyperparameters without systematic search strategies, which compromises the generalizability and robustness of the models across different crops or regional networks. These limitations highlight the urgent need for a dynamic graph neural network framework that can model the temporal evolution of trade structures, enhance structural stability, and support automated hyperparameter optimization—thereby enabling more realistic and accurate modeling of global food trade network dynamics.

To systematically address key challenges in modeling international food trade networks—such as high structural sparsity, temporal complexity, and limited model robustness—this study proposes a novel dynamic variational graph neural network model, IVGAE-TAMA-BO (Improved Variational Graph Autoencoder with Trade-Aware Momentum Aggregation and Bayesian Optimization). Built upon the original static IVGAE framework[31], the proposed model introduces several methodological enhancements across three key dimensions. First, to capture the evolving structural patterns in food trade networks over time, a dynamic graph modeling mechanism is incorporated. By constructing sequential graph inputs across multiple time steps, the model enables end-to-end learning of temporal evolution, thereby improving its ability to predict future trade links. Second, to mitigate the effects of structural sparsity—such as disrupted information propagation and weakened long-term dependency modeling—a novel temporal aggregator with structural memory, termed TAMA, is introduced. This module integrates a unidirectional GRU-based temporal encoder with an exponentially weighted momentum memory mechanism, enabling the model to capture both short-term fluctuations and long-term structural inertia. As a result, it enhances the model's resilience and structural awareness in sparse and unstable networks. Finally, to improve adaptability and performance stability across diverse crop-specific trade networks, Bayesian optimization is employed to automatically search for critical hyperparameters—including learning rate, latent dimensionality, KL divergence weight, and momentum coefficient. This approach overcomes the limitations of manual tuning and significantly enhances the model's generalizability and consistency across varying network topologies.

The key contributions of this study are summarized as follows:

- This study introduces dynamic graph neural networks into the context of international food trade, enabling accurate link prediction for future trade flows between countries.
- We extend the proposed IVGAE framework to handle dynamic graph structures, enabling temporal modeling of evolving trade networks across multiple time steps.

- A novel structural memory module, the Trade-Aware Momentum Aggregator (TAMA), is proposed to enhance structural stability by incorporating learnable momentum, specifically tailored to dynamic and sparse food trade networks.
- To improve predictive performance and generalizability across diverse crop-specific trade networks, Bayesian optimization is employed for automatic hyperparameter tuning.

The remainder of this paper is structured as follows. Section 2 introduces the proposed IVGAE-TAMA-BO framework, detailing its architecture and key components. Section 3 presents the experimental setup and results. Section 4 discusses the main findings, implications, and limitations and future directions. Finally, Section 5 concludes the study.

# 2 Methods

## 2.1 Overview

In this work, we propose IVGAE-TAMA-BO, a dynamic link prediction model designed for evolving global food trade networks. Building upon the IVGAE model[31], our approach incorporates temporal dynamics, momentum-based structural memory, and automated hyperparameter optimization to improve predictive accuracy and generalization.

The model consists of three main components:

- **Improved Variational Graph Autoencoder (IVGAE)[31]**: encodes node representations by combining local structural information and non-local trade-weighted signals from the network.
- **Trade-Aware Momentum Aggregator (TAMA)**: captures temporal patterns in the latent space using a GRU and a momentum-based smoothing mechanism.
- **Bayesian Optimization (BO)[34]**: searches for optimal hyperparameters such as latent dimension, learning rate, and KL weight to enhance model performance across diverse crop-specific trade networks.

At each time step, node features and trade matrices are encoded into latent embeddings using the IVGAE encoder; these embeddings are then aggregated across time by the TAMA module to capture temporal dependencies, and the overall model is optimized via Bayesian Optimization to automatically select hyperparameters that maximize link prediction performance. The overall architecture of the IVGAE-TAMA-BO model is shown in Figure1.

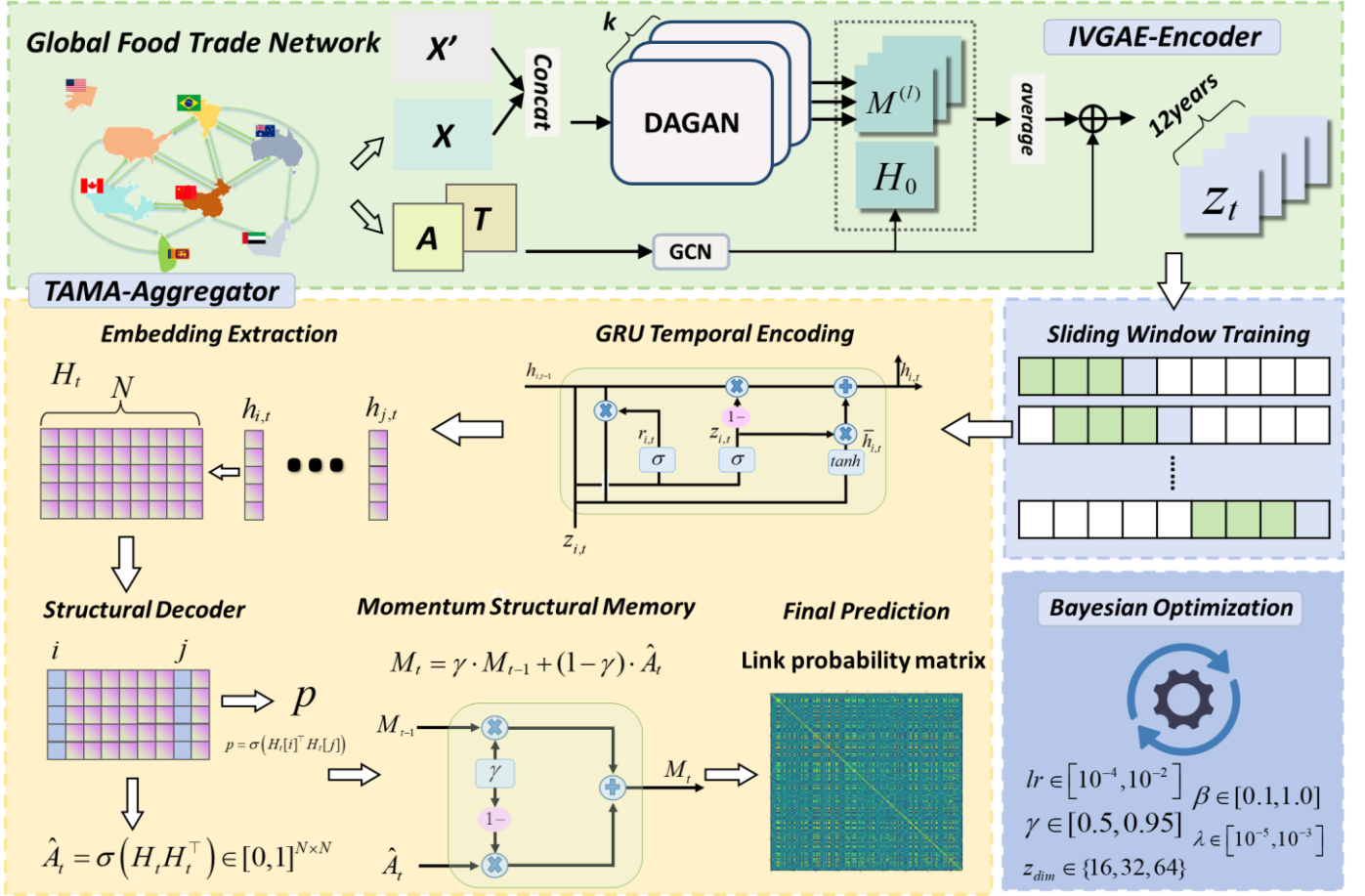

**Figure 1 | IVGAE-TAMA-BO model architecture**

## 2.2 Improved Variational Graph Autoencoder (IVGAE) Encoder

To generate informative and uncertainty-aware node embeddings from evolving grain trade networks, we adopt the IVGAE encoder design originally proposed in IVGAE-GA[31]. The IVGAE framework has been demonstrated to outperform other variational graph models in previous studies, providing a strong foundation for our dynamic extension. This encoder extracts both non-local structural dependencies and local topological features, guided by the trade flow matrix and adjacency structure. At each time step, it acts as the first stage of our dynamic model, encoding input node and edge information into latent representations $Z_t$ under a variational distribution.

### 2.2.1 Input construction

Let $X \in R^{N \times d}$ denote the node feature matrix at a given time step, $A \in \{0,1\}^{N \times N}$ the binary adjacency matrix representing trade existence, and $T \in R^{N \times N}$ the weighted trade flow matrix. To enrich node expressiveness and account for missing or unobservable attributes, we introduce a learnable node embedding matrix $X_p \in R^{N \times d}$, optimized along with the rest of the model. The final input to the encoder is the row-wise concatenation:

$$\tilde{X} = Concat(X, X_p) \tag{1}$$

which serves as the input to both non-local and local feature extractors.

### 2.2.2 Non-local representation via DAGAN modules

To capture high-order and distant relational patterns that may not be accessible through shallow convolution, we employ $k$ parallel Dynamic Adaptive Graph Attention Network (DAGAN) modules, each applying transformation, normalization, and attention-based aggregation. The processing steps are:

**(i) Feature transformation and normalization.** Each DAGAN begins by projecting the input features to a hidden space:

$$M^{(0)} = MLP(\tilde{X}) \tag{2}$$

which is then $\ell 2$-normalized to mitigate magnitude collapse in isolated or low-degree nodes:

$$M^{(0)} = s \cdot \frac{M^{(0)}}{\|M^{(0)}\|_2} \tag{3}$$

where $s$ is a learnable scaling constant, $\|\cdot\|_2$ is $\ell_2$-norm.

**(ii) DropEdge regularization and attention construction.** To avoid over-reliance on fixed graph topology and enhance generalization, we adopt DropEdge[35], randomly removing a fraction, randomly removing a fraction $p$ of edges:

$$A^{\text{drop}} = A \odot M, \quad P^{\text{drop}} = softmax\left(T \odot A^{\text{drop}}\right) \tag{4}$$

where $\odot$ denotes element-wise multiplication and softmax is column-wise.

**(iii) Residual adaptive aggregation.** Each DAGAN performs $L$ iterations of feature propagation:

$$M^{(l)} = \alpha^{(l-1)} \cdot P^{\text{drop}} \cdot M^{(l-1)} + \beta^{(l-1)} \cdot M^{(0)}, \quad l = 1, \ldots, L \tag{5}$$

where $\alpha, \beta \in (0,1)$ are trainable residual weights. This allows the model to adaptively blend initial and aggregated signals layer-by-layer.

### 2.2.3 Local representation via GCN

To complement global attention with localized structural encoding, we apply a two-layer GCN[36] module to the same input:

$$H_0 = GCN(\tilde{X}, \hat{A}), \quad \hat{A} = \hat{D}^{-1/2}(A+I)\hat{D}^{-1/2} \tag{6}$$

where $A \in R_{N \times N}$ is the binary adjacency matrix at a given time step, $I$ is the identity matrix(add self-loops), and $\hat{D}$ is the degree matrix of $A+I$, $\hat{A}$ is the symmetrically normalized adjacency matrix.

### 2.2.4 Feature fusion and variational inference

We fuse the outputs of the GCN and DAGAN modules using average feature fusion, followed by a residual skip connection:

$$M_{\text{all}} = Average\left(\left[H_0, M_1^{(L)}, \ldots, M_k^{(L)}\right]\right), \quad H = M_{\text{all}} + H_0 \tag{7}$$

This step aggregates both global and local signals while preserving the original neighborhood influence. The fused representation $H$ is then passed through two GCN-based heads to obtain the parameters of the variational distribution:

$$\mu = GCN_\mu(ReLU(GCN(H))), \quad \log\sigma = GCN_{\log\sigma}(ReLU(GCN(H))) \tag{8}$$

Where $GCN(\cdot)$ donates a graph convolution layer with shared or independent weights, $ReLU(\cdot)$ is the rectified linear unit activation function, $GCN_\mu$ and $GCN_{\log\sigma}$ are two separate GCN layers used to compute the mean $\mu \in R^{N \times d_z}$ and log-variance $\log\sigma \in R^{N \times d_z}$ of the latent distribution and $d_z$ is the dimension of the latent space. Then, we sample latent embeddings via reparameterization:

$$z = \mu + \sigma \odot \epsilon, \quad \epsilon \sim \mathcal{N}(0, I) \tag{9}$$

To regularize the posterior distribution $q(z \mid X, A)$, we apply a standard Gaussian prior and compute the KL divergence:

$$\mathcal{L}_{KL} = -\frac{1}{2} \sum_{i=1}^{N} \left(1 + 2\log\sigma_i - \mu_i^2 - e^{2\log\sigma_i}\right) \tag{10}$$

This encoder module provides a rich latent representation $Z_t \in R^{N \times d_z}$ that serves as the input for the temporal aggregation module.

## 2.3 Trade-Aware Momentum Aggregator (TAMA)

Modeling the temporal evolution of international food trade networks requires capturing both short-term fluctuations and long-term dependencies in trade structures. To address this, we design the TAMA module, which performs temporal aggregation over a sequence of latent node embeddings using a unidirectional Gated Recurrent Unit (GRU) and a momentum-based structural memory.

The goal of TAMA is to transform a sequence of encoded node representations $\{Z_t\}_{t=1}^{T}$ (produced by the IVGAE encoder) into a temporally enriched embedding $H_T$ and a prediction matrix for future links.

### 2.3.1 Input: sequence of latent representations

At each time step $t$, the IVGAE encoder generates a latent node embedding $z_t \in R^{N \times d}$, where $N$ is the number of nodes (countries) and $d$ is the latent dimension. These embeddings capture the structural and trade-based characteristics of the food trade network at time $t$. We collect the embeddings from the most recent $T$ time steps into a 3D tensor:

$$Z = [z_1, z_2, \ldots, z_T] \in R^{T \times N \times d} \tag{11}$$

This temporal sequence forms the input to TAMA, enabling it to reason over historical dynamics.

### 2.3.2 Temporal encoding via GRU

To capture temporal dependencies in evolving trade networks, we adopt a unidirectional Gated Recurrent Unit (GRU)[37], which processes the sequence of latent node embeddings in the forward direction allowing the model to summarize historical trade dynamics and extract temporal patterns that support future link prediction. The unidirectional GRU provides a balance between temporal modeling capability and computational efficiency. It outputs context-aware hidden states for each node at each time step, which are used for downstream prediction. The GRU treats each node as a separate sequence (along the temporal dimension) and outputs:

$$h_{i,t} = \mathrm{GRU}\left(\{z_{i,t-W+1},\ldots,z_{i,t}\}\right) \qquad (12)$$

where $z_{i,t} \in R^d$ is the latent embedding of node $i$ at time $t$, and $h_{i,t} \in R^d$ is the GRU hidden state output corresponding to the final time step, the GRU operates independently on the temporal sequence of each node over a sliding window of length $W$. Each output $h_{i,t}$ summarizes the historical trade dynamics within the window. The GRU neuron are shown in Figure 2.

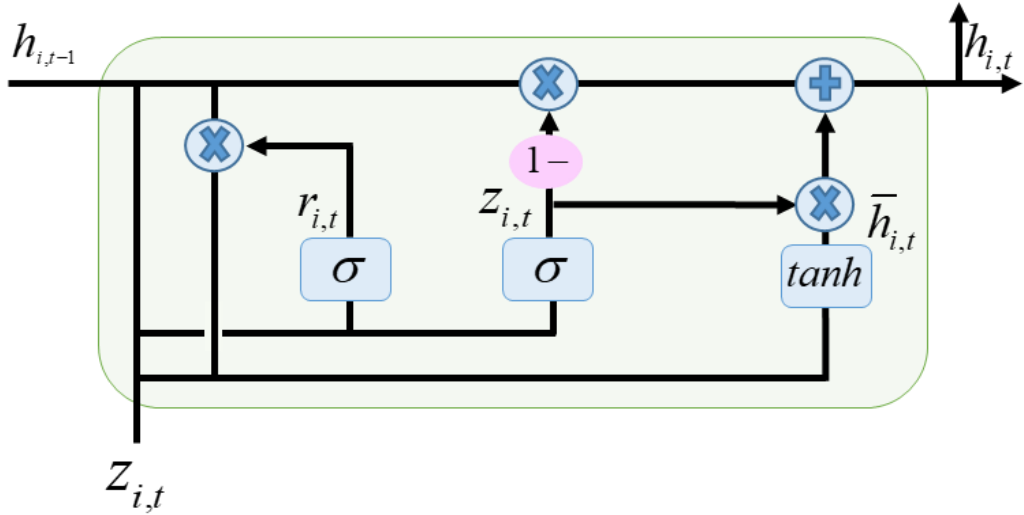

**Figure 2 | GRU neuron.**

### 2.3.3 Embedding extraction for prediction

To better model the dynamic patterns of international food trade, it is crucial to extract temporally enriched node representations from the GRU module. After processing each node's sequence of latent embeddings, the GRU outputs a time-aware hidden state $h_{i,t}$ at each time step $t$, where $d$ denotes the latent dimension. To prepare for downstream prediction, we collect the GRU outputs of all nodes at the final time step into a matrix $H_t$, which represents the temporally smoothed embedding of the entire trade network at time $t$. Each row of $H_t$ corresponds to the temporal embedding of a specific node and captures its historical trade behavior within the input sequence. In our framework, we utilize only the final embedding $H_t$ from the last time step, as it encapsulates sufficient temporal dependencies accumulated across the sequence window to support accurate link prediction.

### 2.3.4 Structural interaction scoring

Following the temporal embedding extraction step, we proceed to estimate the likelihood of future trade connections using a structural interaction scoring mechanism. Specifically, we compute pairwise interaction scores between countries based on the inner product of their final temporal embeddings. The score between country $i$ and country $j$ is computed as:

$$\hat{A}_t[i,j] = \sigma\left(H_t[i]^\top H_t[j]\right) \tag{13}$$

which yields a symmetric matrix of predicted link probabilities:

$$\hat{A}_t = \sigma\left(H_t H_t^\top\right) \in [0,1]^{N \times N} \tag{14}$$

where $\sigma(\cdot)$ is the sigmoid function applied element-wise. This scoring mechanism reflects the model's confidence in the presence of future trade links between country pairs, based on the temporally aggregated and dimensionally aligned node embeddings. Such a design ensures that both sequential dependencies and structural interaction patterns are leveraged in forecasting the evolution of the global food trade network.

**2.3.5 Momentum-based structural memory**

In global food trade networks, while short-term fluctuations and policy shocks can cause temporary changes, the overall trade structure tends to exhibit strong long-term stability and structural inertia—e.g., persistent export-import relationships between key countries[28,30]. To effectively model this temporal persistence, we introduce a Momentum-Based Structural Memory mechanism, designed to retain accumulated knowledge of past trade structures and enhance the model's ability to capture global evolutionary patterns over time.

The design of the momentum-based structural memory in this study is inspired by the classical momentum mechanism widely used in optimization and signal processing[38,39]. In gradient-based optimization, the Polyak momentum update[38] can be expressed as

$$v_t = \beta v_{t-1} + (1-\beta)g_t \tag{15}$$

where $g_t$ denotes the gradient at iteration $t$, $v_t$ is the accumulated velocity term, and $\beta \in (0,1)$ is the momentum coefficient controlling the trade-off between the current update and the historical trend. From a signal processing perspective, this formulation is equivalent to an exponential moving average (EMA)[39,40]:

$$v_t = (1-\beta)\sum_{k=0}^{t-1}\beta^k g_{t-k} \tag{16}$$

which effectively filters out high-frequency noise while preserving low-frequency, long-term trends. Motivated by this property, we adapt the EMA concept from optimization to the domain of dynamic graph structure modeling, replacing the gradient signal $g_t$ with the predicted adjacency score matrix $\hat{A}_t$ at time $t$. Let $M_t \in R^{N \times N}$ donate the denote the structural memory matrix that stores a smoothed representation of historical trade structures. The memory is updated recursively as:

$$M_t = \gamma \cdot M_{t-1} + (1-\gamma) \cdot \hat{A}_t \tag{17}$$

where: $\gamma \in (0,1)$ is a learnable momentum coefficient (initialized near 0.8), which controls the trade-off between historical memory and current prediction, $\hat{A}_t$ is the predicted adjacency score at time $t$, $M_0$ is

initialized as a zero matrix. This recursion can be unfolded into:

$$M_t = (1-\gamma)\sum_{k=0}^{t-1}\gamma^k \hat{A}_{t-k} \qquad (18)$$

which indicating that more recent predictions receive higher weights, while older structures decay exponentially at rate $\gamma^k$. These formulation effectively acts as an exponential moving average over the sequence of predicted trade structures, smoothing out noisy short-term variations while preserving long-term consistency. By recursively blending past predictions with current outputs, the momentum memory enhances robustness and prevents erratic structural shifts, which is especially beneficial in sparse or highly dynamic food trade networks. Additionally, this memory matrix serves as a stable structural prior for subsequent modules, such as the residual attention propagation, making it a key architectural innovation in our dynamic graph modeling framework.

**2.3.6 Final prediction**

The final output of TAMA is a fused link prediction matrix $A_{logits}$, combining instantaneous structural inference and long-term memory:

$$A_{\text{logits}} = H_T H_T^\top + \beta \cdot M_T \qquad (19)$$

where, $\beta \in R$ is another learnable scalar parameter that balances the influence of the momentum-enhanced structure. A higher $\beta$ enforces more historical persistence, while a lower $\beta$ focuses on the most recent dynamics.

In summary, TAMA fuses recurrent sequence modeling and exponential memory to produce stable, temporally-aware link predictions. It enables the model to adapt to rapid trade changes while preserving structural continuity — a critical property for forecasting in dynamic and sparse global trade networks.

## 2.4 Loss function

The training objective consists of two components: a reconstruction loss and a regularization term from variational inference. Given the predicted adjacency logits $A_{logits}$ and the ground-truth future adjacency matrix $A_{true}$, the total loss is defined as:

$$\mathcal{L} = \underbrace{\text{BCE}(A_{\text{logits}}, A^{\text{true}})}_{\text{Reconstruction loss}} + \lambda \cdot \underbrace{\text{KL}(q(z|\ X, A, T) \| p(z))}_{\text{KL divergence}} \qquad (20)$$

where, $\text{BCE}(\cdot,\cdot)$ is the binary cross-entropy loss that measures the discrepancy between the predicted adjacency logits and the ground-truth link labels, $A_{\text{logits}}$ and $A^{\text{true}}$ denote the predicted and ground-truth adjacency matrices, respectively, $KL(q(z|\ X, A, T) \| p(z))$ is the Kullback–Leibler divergence between the approximate posterior and the standard Gaussian prior, and $\lambda$ is a tunable weight balancing accuracy and regularization.

## 2.5 Bayesian optimization for hyperparameter search

To maximize predictive performance and avoid manual hyperparameter tuning, we adopt Bayesian Optimization (BO)[34] to automatically search for the optimal configuration of the proposed IVGAE-TAMA model. BO provides an efficient, sample-efficient strategy to explore the hyperparameter space by modeling the objective function (e.g., final AUC) as a probabilistic surrogate and iteratively selecting promising candidates via acquisition functions.

Specifically, we optimize the following key hyperparameters: learning rate $\in \left[10^{-4}, 10^{-2}\right]$ (log scale), latent dimension $z_{dim} \in \{16, 32, 64\}$, momentum coefficient $\gamma \in [0.5, 0.95]$, structural memory weight $\beta \in [0.1, 1.0]$, KL divergence weight $\lambda \in \left[10^{-5}, 10^{-3}\right]$. Although $\gamma$ and $\beta$ are learnable parameters, we use Bayesian Optimization to search for their optimal initial values, which significantly affects model stability and final performance, especially in sparse and dynamic graph settings.

During training, BO iteratively evaluates 100 candidate configurations by training the model and recording the validation AUC. Poor candidates are pruned early via a median-based early stopping policy. The final selected configuration is used to retrain the model from scratch to obtain the reported performance. This automated tuning process significantly improves robustness and reproducibility across different crops and temporal horizons, while reducing the need for manual trial-and-error.

## 2.6 Sliding window training strategy

To capture temporal evolution in food trade networks, we adopt a sliding window training strategy[41] that transforms sequential data into overlapping temporal segments. Let the temporal food trade data be represented as a sequence of graph snapshots $\mathcal{G}$:

$$\mathcal{G} = \{(X_1, A_1, T_1), (X_2, A_2, T_2), \ldots, (X_S, A_S, T_S)\} \tag{21}$$

where, $X_t \in R^{N \times F}$ represents node feature matrix at time $t$, $A_t \in \{0,1\}^{N \times N}$ is the binary adjacency matrix, $X_t \in R^{N \times N}$ donates trade volumes, and $S$ is the total number of time steps. we construct training samples using a fixed window length $w$. Each training instance contains: node features $\{X_t, \ldots, X_{t+w-1}\}$, adjacency matrices $\{A_t, \ldots, A_{t+w-1}\}$, trade weights $\{T_t, \ldots, T_{t+w-1}\}$, and target $A_{t+w}$. The model is trained to predict the future link structure $A_{t+w}$ based on the past $w$ time steps.

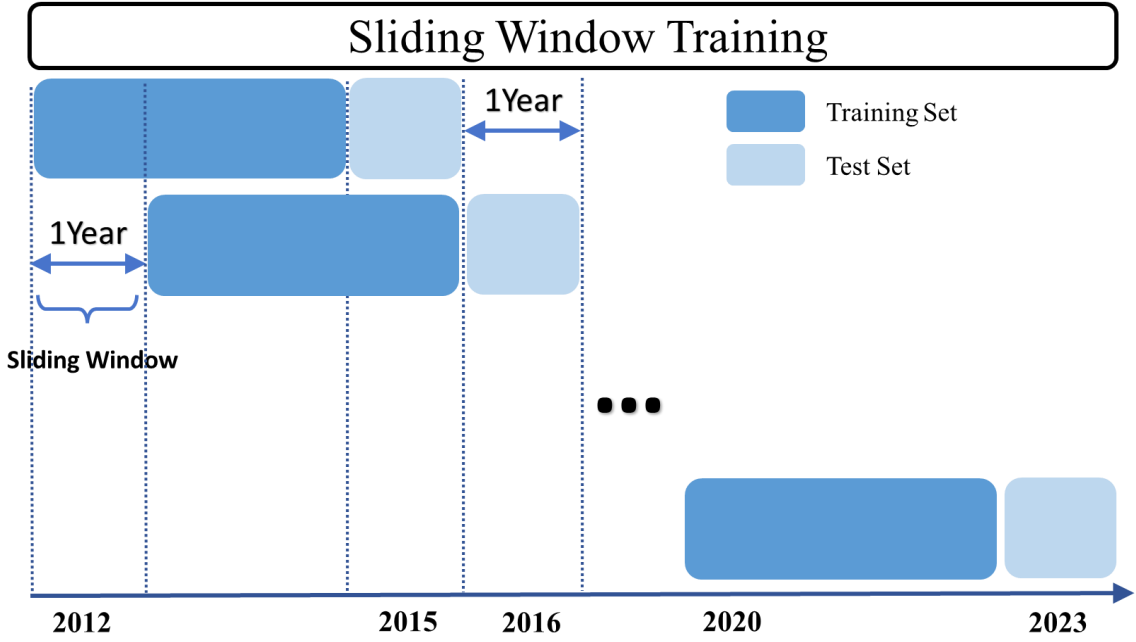

**Figure 3 | Framework of the sliding window training strategy**

## 3 Experiments and Results

### 3.1 Experimental setup

To evaluate the effectiveness and robustness of our proposed IVGAE-TAMA-BO model for link prediction in global grain trade networks, we design a comprehensive experimental setup as follows:

**3.1.1 Datasets and preprocessing**

Our experiments are conducted on five major grain categories: barley, corn, rice, soya beans, and wheat, covering the years 2012 to 2023. For each crop, we construct a dynamic directed graph sequence where nodes represent countries and edges represent annual bilateral trade flows.

At each time step $t$, the data includes: A binary adjacency matrix $A_t \in \{0,1\}^{N \times N}$, where $A_t(i,j) = 1$ indicates that country $i$ exported the crop to country $j$ in year $t$; A weighted trade matrix $T_t \in R^{N \times N}$, capturing the actual trade volume (in metric tons) between exporters and importers; A node feature matrix $X_t \in R^{N \times 4}$, where each node is described by four standardized country-level attributes: GDP, agricultural employment ratio, population, and grain production, retrieved from the World Bank[42]. All bilateral trade matrices $\{A_t, T_t\}$ are obtained from the FAO Food Trade Database[43], while the node-level indicators are sourced from the World Bank Open Data platform[42]. To ensure consistency across years and sources, country identities are unified using ISO country codes.

During preprocessing, missing values in the World Bank indicators are filled via linear interpolation over time. To eliminate scale bias and support stable learning dynamics, all node features are then normalized per year to zero mean and unit variance. This preprocessing pipeline results in a temporally aligned, multi-modal dataset that captures both the topological evolution of grain trade networks and the structural heterogeneity of countries.

**3.1.2 Experimental environment**

The experiments are conducted on a high-performance local machine; detailed hardware and software configurations are listed in Table 1.

**Table 1 | Experimental environment**

| Development Environment | Experimental Configuration |
| --- | --- |
| CPU | i9-14900HX |
| GPU | NVIDIA GeForce RTX 4070 Laptop GPU |
| Memory | 32GB |
| OS | Windows 11 64-bit |
| Python Version | 3.12.9 |
| Deep learning framework | PyTorch 2.5.1+cuda11.8 |
| Programming tools | Anaconda |
| IDE | PyCharm |

**3.1.3 Evaluation indicators**

Following common evaluation practices in mainstream link prediction models, we adopt two widely used metrics to quantitatively assess the performance of our model:

**Area under the ROC curve (AUC)**:

AUC measures the ability of the model to distinguish between existing and non-existing links. Given a set of positive edges $\mathcal{E}^+$ and an equal-sized set of negative edges $\mathcal{E}^-$, AUC estimates the probability that a randomly chosen positive edge has a higher predicted score than a randomly chosen negative one:

$$AUC = \frac{1}{|\mathcal{E}^+||\mathcal{E}^-|} \sum_{(i,j)\in\mathcal{E}^+} \sum_{(u,v)\in\mathcal{E}^-} \mathbb{I}(s_{ij} > s_{uv}) \quad (22)$$

where $s_{ij}$ is the predicted score (e.g., $A_{\text{logits}}[i,j]$), and $\mathbb{I}(\cdot)$ is the indicator function. A higher AUC indicates better global discriminative capability.

**Average precision (AP)**:

AP summarizes the precision-recall curve as the weighted mean of precisions at each threshold, emphasizing the model's performance on correctly ranking positive links. AP summarizes the precision-recall curve and is defined as:

$$AP = \sum_{k=1}^{n} P(k) \cdot \Delta r(k) \quad (23)$$

where $P(k)$ is the precision at cutoff $k$, and $\Delta r(k)$ is the change in recall. This metric emphasizes correct ranking of positive links among all predictions. It is particularly useful in sparse graphs where the number of negative links far outweighs the positives.

Both metrics are calculated over the upper triangular part of the predicted adjacency matrix, excluding self-loops.

## 3.2 Comparison with baseline models

To evaluate the effectiveness of our proposed IVGAE-TAMA model, we conduct comprehensive comparative experiments against a diverse set of baseline methods. All models are trained under consistent experimental settings using the same set of hyperparameters to ensure fairness, with detailed configurations shown in Table 2. Each model is trained over 300 epochs and repeated 10 times with different random seeds, and the average performance is reported to mitigate the effects of stochastic fluctuations. Each training sample corresponds to a full-graph snapshot of trade interactions for a specific year, and temporal input is constructed using a sliding window of four consecutive years. The final evaluation is conducted on the last sliding window.

**Table 2 | Fixed training configuration for optimal hyperparameter runs**

| Configuration Item | Value |
| --- | --- |
| Epochs | 300 |
| Optimizer | Adam[40] |
| Learning Rate | 0.001 |
| Kl Weight | 0.0001 |
| Batch Mode | Full-graph |
| DAGAN Dropout Rate (p) | 0.2 |
| DAGAN Layers (L) | 2 |
| DAGAN Heads (k) | 3 |
| Sliding Window Size | 4 |
| Repeats (Seeds) | 10 (1000 to 1009) |

The baselines are selected to represent three distinct categories of modeling approaches for dynamic link prediction. First, we include the original IVGAE model, which operates on a static graph constructed from the year 2023. This model serves as a strong variational encoder baseline but lacks temporal modeling capabilities, thus providing a lower bound reference. Second, we incorporate several IVGAE variants that differ in their temporal aggregation strategies, including GRU[37], LSTM[44], TCN[45], Attention[45], Transformer[45]. These models share the same variational encoder as IVGAE but apply different recurrent or convolutional modules to capture temporal dependencies in latent representations. Finally, we compare with representative dynamic graph neural networks, including DyGCN[46], GCRN[47], TGCN[48], and two versions of EvolveGCN (H and O)[49], which are designed to operate directly on evolving graph sequences. These models provide complementary perspectives by learning temporal patterns from raw graph snapshots, rather than variational latent spaces. Since

IVGAE has already demonstrated superior performance over other variational graph models in the original study[31], we do not include additional variational graph autoencoder baselines for comparison.

The visualization of AUC and AP results is shown in Figure 4, with the corresponding numerical values summarized in Tables 4 and 5, respectively. As shown in Figure 4A and detailed in Tables 4, IVGAE-TAMA achieves AUC scores exceeding 94% on barley, rice, soya beans, and wheat, and maintains strong performance above 93% on the remaining crops, ranking first on four out of five crops, demonstrating its strong and reliable link prediction capability. As shown in Figure 4B and detailed in Tables 5, its AP scores are likewise competitive, ranking first on three out of five crops. Although it is marginally outperformed by IVGAE-GRU and IVGAE-LSTM on rice (by approximately 1%) and on wheat (by only 0.2%), it consistently remains among the top-performing models across all crop-specific trade networks. These results highlight the model's strong generalization capability and robustness across diverse commodity-specific trade networks.

In contrast to these strong performances, the static IVGAE model consistently underperforms across all crops and evaluation metrics. For instance, its AUC scores range from 77.92% (rice) to 87.51% (barley), and AP scores range from 81.65% (rice) to 89.92% (barley), which are significantly lower than those achieved by dynamic models. In comparison, IVGAE-based dynamic variants such as IVGAE-GRU, IVGAE-LSTM, and IVGAE-TAMA generally achieve AUC and AP values above 93%, with IVGAE-TAMA reaching up to 96.55% AUC and 96.58% AP on barley. General dynamic GNNs (e.g., DyGCN, GCRN) also outperform the static baseline, although to a lesser extent, typically yielding AUC/AP in the 84%–91% range. These results suggest that incorporating temporal modeling substantially improves predictive performance by better capturing the evolving nature of trade relationships. Notably, IVGAE-based dynamic variants outperform general dynamic GNNs by a margin of 5%–10% in both AUC and AP on most crops, underscoring the advantages of combining variational inference with tailored temporal aggregation for dynamic link prediction in food trade networks.

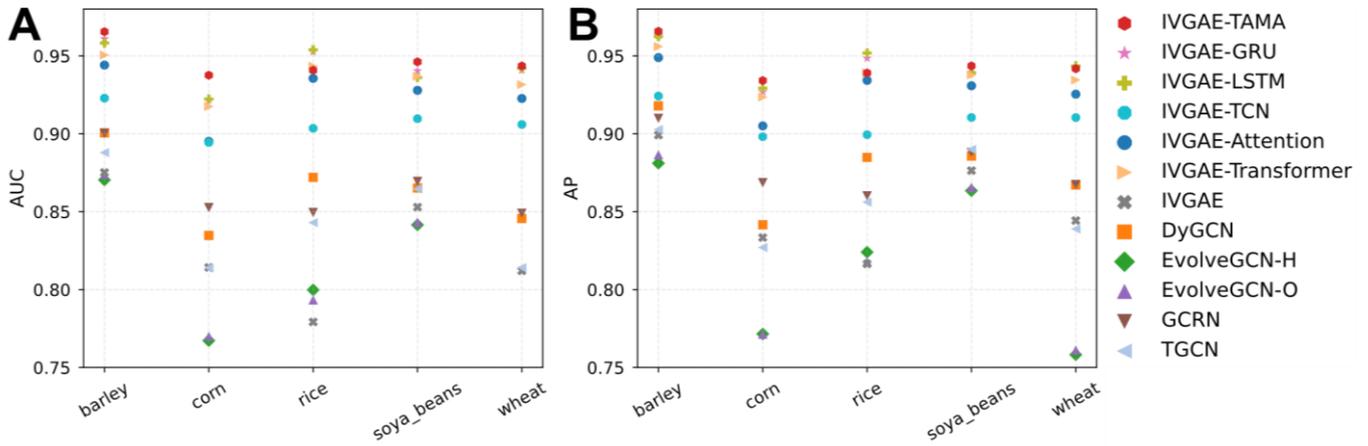

**Figure 4 | Model comparison on five crops datasets. (A)** Area Under the ROC Curve (AUC) and **(B)** Average Precision (AP) for different dynamic graph models.

Table 3. AUC performance (%) of different dynamic graph models on five crops datasets

| Model | Barley | Corn | Rice | Soya beans | Wheat |
| --- | --- | --- | --- | --- | --- |
| IVGAE | 87.51 ± 3.14 | 81.42 ± 2.62 | 77.92 ± 4.64 | 85.29 ± 3.22 | 81.21 ± 4.10 |
| IVGAE-GRU | 96.10 ± 0.41 | 92.20 ± 0.34 | 95.27 ± 0.44 | 94.05 ± 0.47 | 94.09 ± 0.46 |
| IVGAE-LSTM | 95.82 ± 0.63 | 92.23 ± 0.40 | **95.39 ± 0.56** | 93.61 ± 0.59 | 94.21 ± 0.29 |
| IVGAE-TCN | 92.29 ± 0.70 | 89.44 ± 0.83 | 90.35 ± 1.25 | 90.97 ± 0.60 | 90.60 ± 0.74 |
| IVGAE-Attention | 94.41 ± 0.52 | 89.51 ± 0.62 | 93.56 ± 0.44 | 92.79 ± 0.69 | 92.27 ± 0.57 |
| IVGAE-Transformer | 95.06 ± 0.59 | 91.75 ± 0.48 | 94.33 ± 0.65 | 93.69 ± 0.60 | 93.16 ± 0.76 |
| DyGCN | 90.06 ± 0.75 | 83.48 ± 1.34 | 87.20 ± 0.97 | 86.52 ± 0.92 | 84.56 ± 1.38 |
| EvolveGCN-H | 87.04 ± 0.63 | 76.73 ± 0.95 | 79.98 ± 0.88 | 84.15 ± 0.98 | 73.26 ± 0.89 |
| EvolveGCN-O | 87.38 ± 0.71 | 76.99 ± 0.92 | 79.33 ± 1.12 | 84.31 ± 0.83 | 73.67 ± 0.86 |
| GCRN | 90.03 ± 1.33 | 85.26 ± 1.76 | 84.95 ± 0.99 | 86.94 ± 0.92 | 84.90 ± 0.83 |
| TGCN | 88.79 ± 0.98 | 81.39 ± 1.47 | 84.30 ± 1.60 | 86.46 ± 0.99 | 81.41 ± 1.47 |
| IVGAE-TAMA (ours) | **96.55 ± 0.38** | **93.76 ± 0.30** | 94.08 ± 0.52 | **94.62 ± 0.37** | **94.35 ± 0.38** |

Table 4. AP performance (%) of different dynamic graph models on five crops datasets

| Model | Barley | Corn | Rice | Soya beans | Wheat |
| --- | --- | --- | --- | --- | --- |
| IVGAE | 89.92 ± 2.29 | 83.34 ± 2.89 | 81.65 ± 4.00 | 87.64 ± 3.74 | 84.43 ± 3.40 |
| IVGAE-GRU | 96.31 ± 0.37 | 92.68 ± 0.37 | 94.85 ± 0.69 | 94.22 ± 0.41 | 94.27 ± 0.43 |
| IVGAE-LSTM | 96.25 ± 0.37 | 92.92 ± 0.37 | **95.18 ± 0.68** | 93.91 ± 0.46 | **94.35 ± 0.27** |
| IVGAE-TCN | 92.42 ± 0.69 | 89.81 ± 0.70 | 89.94 ± 1.12 | 91.04 ± 0.49 | 91.04 ± 0.64 |
| IVGAE-Attention | 94.89 ± 0.43 | 90.51 ± 0.33 | 93.43 ± 0.48 | 93.08 ± 0.47 | 92.54 ± 0.66 |
| IVGAE-Transformer | 95.60 ± 0.42 | 92.35 ± 0.25 | 93.91 ± 0.54 | 93.77 ± 0.53 | 93.45 ± 0.56 |
| DyGCN | 91.79 ± 0.78 | 84.16 ± 1.42 | 88.49 ± 1.29 | 88.56 ± 0.66 | 86.72 ± 0.94 |
| EvolveGCN-H | 88.11 ± 0.80 | 77.15 ± 1.06 | 82.40 ± 1.43 | 86.35 ± 0.80 | 75.82 ± 1.41 |
| EvolveGCN-O | 88.65 ± 0.82 | 77.14 ± 1.24 | 81.80 ± 1.60 | 86.56 ± 0.87 | 76.11 ± 1.32 |
| GCRN | 90.99 ± 1.58 | 86.86 ± 1.40 | 86.02 ± 1.42 | 88.80 ± 0.85 | 86.74 ± 1.04 |
| TGCN | 90.27 ± 0.73 | 82.70 ± 1.13 | 85.61 ± 1.44 | 89.00 ± 0.60 | 83.89 ± 1.36 |
| IVGAE-TAMA (ours) | **96.58 ± 0.38** | **93.42 ± 0.32** | 93.90 ± 0.33 | **94.35 ± 0.50** | 94.17 ± 0.45 |

## 3.3 Effectiveness of bayesian hyperparameter optimization

To ensure robust and consistent performance across different crop-specific grain trade networks, the proposed IVGAE-TAMA model employs Bayesian Optimization (BO), implemented via Optuna, to automatically tune key hyperparameters such as the learning rate, dropout rate, GRU hidden dimensions, and trade-aware momentum factors. This automated search strategy enables the model to identify optimal configurations that enhance predictive accuracy and generalizability without relying on manual tuning.

For each crop, Bayesian optimization is conducted with 100 trials, where each trial corresponds to a unique hyperparameter configuration, and each configuration is trained and evaluated 10 times using different random seeds (from 1000 to 1009), exploring key hyperparameters such as learning rate, latent dimension, KL divergence weight, and momentum coefficients, and the mean and standard deviation of the evaluation metrics are reported. After identifying the best configuration, we retrain the model using the selected parameters and repeat the training 10 times under different random seeds (from 1000 to 1009) to report the mean and standard deviation of evaluation metrics. As a comparison, we also train the IVGAE-BiTAMA model using the fixed hyperparameter settings listed in Table 2, while keeping the number of training epochs consistent at 500 to ensure fairness across experimental conditions.

Figure 5 demonstrates the training convergence curves of IVGAE-TAMA-BO across five crops. Figure 5A depicts the evolution of the training loss over 500 epochs for each crop. All curves exhibit a rapid decrease in the initial 0–200 epochs, indicating effective convergence behavior. As training progresses, the loss flattens and stabilizes, with minor fluctuations, reflecting the model's ability to fit the dynamic food trade networks while avoiding severe overfitting. Notably, barley and soya beans achieve the lowest final losses, suggesting relatively easier link prediction tasks or better signal quality for these crops. In contrast, rice and wheat maintain higher loss values, potentially due to greater noise, sparsity, or instability in their trade patterns. Figure 5B illustrates the evolution of the AUC metric during training. A sharp increase in AUC values during the first 50 epochs across all crops, indicating rapid improvement in the model's classification ability between existent and non-existent trade links. Final AUC values consistently exceed 0.95 for all crops, with barley reaching the highest score (~0.98), suggesting strong discriminative capacity of the model. The convergence is notably smooth and stable after epoch 200, confirming the robustness and reliability of the proposed model. Figure 5C shows the Average Precision metric across epochs. Similar to AUC, AP exhibits a steep rise in early epochs and stabilizes around 0.95–0.98. The barley curves dominate, achieving AP values close to 0.98, consistent with its low training losses. The slight oscillations observed in curves reflect minor volatility in ranking precision but remain within acceptable bounds. The optimized hyperparameters for each crop after Bayesian tuning are summarized in Table 5.

As shown in Tables 6 and 7, IVGAE-TAMA-BO consistently outperforms the baseline IVGAE-TAMA across all five crops in both AUC and AP metrics, demonstrating the effectiveness of Bayesian hyperparameter optimization. In terms of AUC, notable improvements are observed on rice (from 93.65% ± 0.73 to 96.11% ±

0.68) and wheat (from 94.67% ± 0.35 to 96.06% ± 0.58), while barley and corn also show steady gains. Similarly, for AP scores, IVGAE-TAMA-BO achieves clear improvements on barley (from 96.61% ± 0.29 to 97.27% ± 0.42) and rice (from 93.62% ± 0.59 to 94.84% ± 0.75), along with consistent enhancements on other crops. These results confirm that optimizing key hyperparameters significantly boosts model performance and robustness across diverse trade network structures.

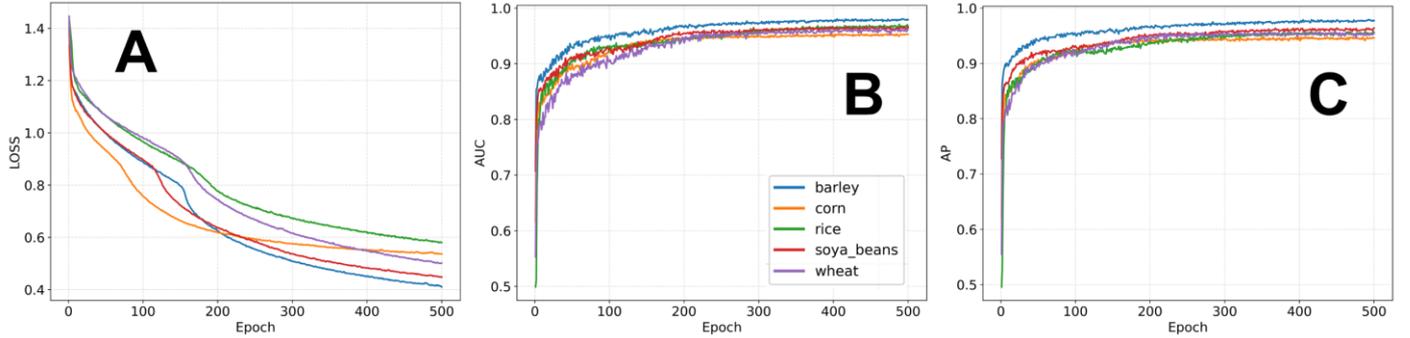

**Figure 5 | Training convergence curves of IVGAE-TAMA-BO across five crops. (a) Training loss; (b) Area under the ROC curve (AUC); (c) Average precision (AP).**

Table 5 | Optimized hyperparameters for each crop

| Crop | Learning rate | $z_{dim}$ | Init $\gamma$ | Init $\beta$ | $\lambda$ |
| --- | --- | --- | --- | --- | --- |
| barley | 0.00110 | 32 | 0.78893 | 0.67116 | 4.98e-4 |
| corn | 0.00139 | 16 | 0.87259 | 0.13965 | 2.78e-4 |
| rice | 0.00075 | 16 | 0.54358 | 0.26047 | 9.79e-4 |
| soya_beans | 0.00119 | 32 | 0.62708 | 0.53546 | 4.52e-4 |
| wheat | 0.00089 | 64 | 0.74063 | 0.63802 | 7.37e-4 |

Table 6 | Comparison of AUC scores (%) for IVGAE-TAMA and IVGAE-TAMA-BO

| Model | Barley | Corn | Rice | Soya beans | Wheat |
| --- | --- | --- | --- | --- | --- |
| IVGAE-TAMA | 96.65 ± 0.31 | 94.34 ± 0.29 | 93.65 ± 0.73 | 94.68 ± 0.39 | 94.67 ± 0.35 |
| IVGAE-TAMA-BO | **97.45 ± 0.38** | **94.97 ± 0.29** | **96.11 ± 0.68** | **95.72 ± 0.53** | **96.06 ± 0.58** |

Table 7 | Comparison of AP scores (%) for IVGAE-TAMA and IVGAE-TAMA-BO

| Model | Barley | Corn | Rice | Soya beans | Wheat |
| --- | --- | --- | --- | --- | --- |
| IVGAE-TAMA | 96.61 ± 0.29 | 93.76 ± 0.40 | 93.62 ± 0.59 | 94.40 ± 0.47 | 94.36 ± 0.44 |
| IVGAE-TAMA-BO | **97.27 ± 0.42** | **94.19 ± 0.47** | **94.84 ± 0.75** | **94.77 ± 0.82** | **95.56 ± 0.57** |

## 3.4 Sensitive analysis of sliding window size

To evaluate the impact of sliding window sequence length on the predictive performance of the IVGAE-TAMA model, we conduct a sensitivity analysis on the window size. The sliding window is used to construct the temporal input sequence for the model, and its length determines the number of historical time steps incorporated in each prediction, thereby influencing the model's ability to capture temporal dependencies. A window that is too short may fail to reflect long-term structural evolution, while an overly long window may introduce redundant or noisy information, potentially impairing the model's generalization capability. Therefore, this experiment aims to systematically analyze how different sequence lengths affect prediction performance and to provide guidance for appropriate parameter selection in practical applications. We evaluate the impact of sliding window sizes ranging from 3 to 8 on model performance. For each window length, the IVGAE-TAMA model is trained using the fixed hyperparameters in Table 2. Each setting is repeated 10 times with random seeds from 1000 to 1009, and the average AUC and AP scores with standard deviations are reported.

Figure 6 presents the impact of sliding window size (ranging from 3 to 8) on the AUC and AP performance of IVGAE-TAMA across five crop-specific trade networks. (A) Barley: Both AUC and AP scores remain high at window sizes 3 and 4 (around 0.965), but gradually decline as the window length increases beyond 5. This suggests that shorter historical sequences are more informative for barley trade prediction. (B) Corn: A clear downward trend is observed in both AUC and AP as the window size increases, with AUC dropping from approximately 0.937 (window=3) to 0.918 (window=8). This indicates the model performs best with shorter windows and may be sensitive to noisy or outdated historical inputs. (C) Rice: Performance remains relatively stable across different window sizes, with minor fluctuations. This implies that rice trade patterns are less sensitive to window length and the model is robust across various temporal input spans. (D) Soya beans: The best performance is observed at window size 4, with AUC and AP reaching approximately 0.945 and 0.947, respectively. Both metrics decline when the window size increases to 5 and 6, but show a slight improvement at sizes 7 and 8, suggesting that the model performance improves again with longer temporal context. (E) Wheat: Both metrics consistently decline as the window size increases, suggesting that recent historical information is more valuable and that longer windows dilute predictive signals. (F) Average trend: The average performance across crops peaks at window sizes 3–4 and decreases with longer windows. This supports the conclusion that shorter window sizes (especially 4) strike a better balance between capturing relevant temporal patterns and avoiding noise accumulation.

The sensitivity analysis demonstrates that a sliding window size of 4 offers the most reliable performance for most crops. Excessively long sequences tend to reduce model accuracy, possibly due to the inclusion of outdated or less relevant structural information. These findings highlight the importance of carefully selecting temporal context length in dynamic graph modeling.

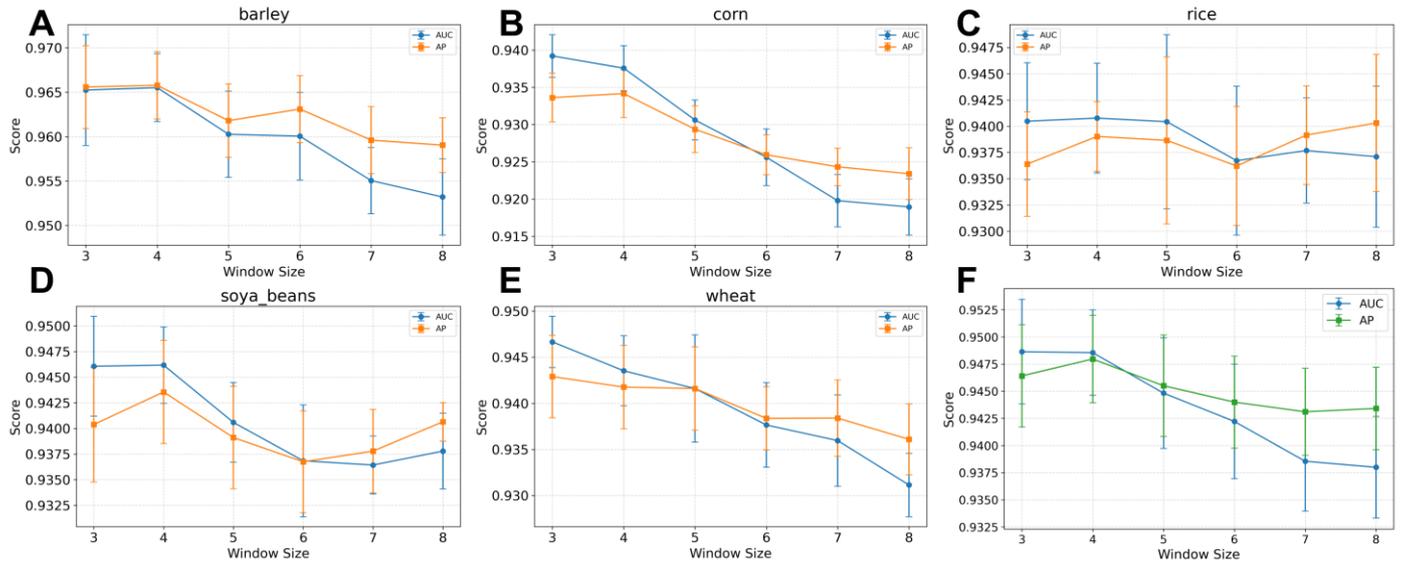

**Figure 6 | Sensitivity analysis of sliding window size on AUC and AP scores across five crops.** (A) barley, (B) corn, (C) rice, (D) soya beans, (E) wheat, (F) average of five crops.

## 4 Discussions

### 4.1 Summary of results

The experimental findings of this study systematically validate the effectiveness and robustness of the proposed IVGAE-TAMA-BO model for dynamic link prediction in global food trade networks. The IVGAE-TAMA model consistently outperforms all baseline approaches across the five major crop-specific datasets, including the static IVGAE model, other IVGAE dynamic variants, and other state-of-the-art dynamic graph neural network (DGNN) models, demonstrating strong generalization capability and predictive accuracy.

In comparative experiments, the IVGAE-TAMA model exhibits substantial effectiveness and robustness in performing dynamic link prediction within global food trade networks. As shown in Tables 3 and 4, IVGAE-TAMA generally achieves the highest AUC and AP scores across the five major crop-specific datasets. For instance, the model attains an AUC of 96.55% and an AP of 96.58% on the barley dataset, and maintains similarly strong performance on rice, soya beans, and wheat, with most scores exceeding 94%. Moreover, IVGAE-TAMA generally outperforms other IVGAE-based dynamic variants such as IVGAE-GRU, IVGAE-LSTM, and IVGAE-Transformer, and maintains a consistent advantage of 5–10% in AUC and AP over general dynamic GNNs like DyGCN, GCRN, and EvolveGCN. These results underscore the model's superior ability to capture evolving trade structures and temporal dependencies. Compared to the static IVGAE baseline, which yields AUC and AP scores below 88 percent and 90 percent respectively across all crop datasets, most dynamic models—including IVGAE-based variants such as TAMA, GRU, LSTM, and Transformer, as well as general dynamic graph neural networks like DyGCN and GCRN—exhibit notable performance improvements. These findings highlight that temporal modeling is more effective in capturing the dynamic information embedded in evolving food trade networks.

A key contributor to this performance is the Trade-Aware Momentum Aggregator (TAMA) module, which integrates GRU-based temporal aggregation with a momentum-driven structural memory mechanism. This hybrid design enables the model to capture both short-term dynamics and long-term structural inertia inherent in food trade systems. Compared to alternative temporal aggregation mechanisms—including GRU, LSTM, Transformer, TCN, and Attention—TAMA consistently achieves superior results. For instance, it records AUC values of 96.55% on barley and 94.62% on soya beans, confirming its enhanced ability to model evolving structural patterns, especially in sparse and unstable networks. The introduction of momentum-based structural memory proves critical for extracting consistent, meaningful representations from temporally evolving graphs.

In addition, the model benefits from Bayesian optimization, which is employed to tune essential hyperparameters such as learning rate, latent dimensionality, KL divergence weight, and momentum coefficients. This automated optimization strategy yields consistent performance gains across all crops. Compared to the fixed-parameter IVGAE-TAMA, the optimized IVGAE-TAMA-BO shows notable improvements, including AUC increases of 2.46% on rice and 1.39% on wheat. These results validate the importance of data-driven, adaptive hyperparameter tuning in dynamic graph modeling. Finally, the sensitivity analysis of sliding window lengths further reveals the impact of temporal input size on model performance, providing valuable guidance for selecting an appropriate temporal granularity in dynamic modeling. A window size of 4 consistently provides the best trade-off between performance and temporal context, whereas shorter windows fail to capture sufficient structural information, and longer windows introduce redundancy and noise. This confirms that a moderate and well-chosen temporal input length is optimal for balancing expressiveness and generalization in dynamic food trade networks.

In summary, IVGAE-TAMA-BO achieves strong predictive performance by integrating variational graph encoding, momentum-based structural memory, temporal sequence modeling, and automated hyperparameter optimization into a unified framework. The model achieves state-of-the-art performance in dynamic food trade link prediction, exhibiting strong robustness and generalization across heterogeneous network structures. Its practical potential is well-suited for critical applications such as international trade monitoring, early warning systems, and global food security assessment.

## 4.2 Methodological innovations

This study introduces several methodological innovations in the field of dynamic graph modeling and international food trade network prediction, substantially enhancing the model's expressive capacity and predictive performance in networks characterized by high sparsity, structural instability, and pronounced temporal evolution.

First, building upon the original static IVGAE model, we extend the variational graph autoencoder (IVGAE) framework to dynamic graph settings for the first time, enabling the modeling of trade network structures across multiple time steps and capturing cross-year structural evolution. This extension not only improves the model's capability to represent dynamic relationships but also broadens the application scope of variational graph neural networks in complex economic systems.

Second, we propose the Trade-Aware Momentum Aggregator (TAMA), which is the core and most innovative component of our framework. Designed to address the high sparsity, temporal volatility, and long-term dependencies of global food trade networks, TAMA combines a unidirectional GRU for capturing short-term structural fluctuations with an exponentially weighted momentum-based memory that preserves persistent trade patterns over time. This hybrid design enables the model to filter out transient noise, maintain stable structural awareness, and mitigate information loss in sparse or perturbed graphs, thereby achieving a balanced representation of both short-term dynamics and long-term structural inertia for robust and accurate link prediction.

Third, to improve robustness and generalization across diverse network structures, we incorporate Bayesian optimization for automated hyperparameter search, covering critical parameters such as latent dimensionality, learning rate, KL divergence weight, and momentum coefficients. This approach not only boosts model performance but also improves the stability and reproducibility of experimental results, avoiding the uncertainty inherent in traditional experience-based hyperparameter tuning.

Finally, we develop a sliding-window-based temporal modeling strategy to standardize the handling of dynamic graph sequence inputs. This strategy efficiently organizes historical graph structure information, allowing the model to learn from multiple consecutive time steps and predict future structures. It achieves a balance between modeling efficiency and temporal context completeness, and offers strong generality and scalability for different temporal granularities and sequence lengths.

In summary, IVGAE-TAMA-BO incorporates substantive improvements in encoder architecture, temporal modeling mechanisms, structural memory design, and optimization strategies, resulting in a stable, efficient, and widely adaptable framework for dynamic graph neural network prediction. This framework provides a powerful tool for modeling the evolution of international food trade networks.

### 4.3 Implications for practice

The IVGAE-TAMA-BO model proposed in this study demonstrates consistently high predictive accuracy and robustness across diverse crop-specific food trade networks, yielding several significant implications for real-world applications.

First, the model's ability to accurately forecast the emergence and dissolution of trade links enables stakeholders—including governments, international organizations, and agricultural policy researchers—to proactively identify risks of structural shifts in global trade, thereby improving preparedness for disruptions due to geopolitical tensions, policy interventions, or supply chain shocks. Second, due to its strong generalization capability in sparse and dynamic network environments, the proposed framework can be readily extended to other domains of international trade, such as energy, fertilizers, and livestock, where evolving structural patterns also require robust dynamic modeling. Moreover, the model's end-to-end design, which jointly integrates structural information and economic node attributes, facilitates practical integration with real-world decision support systems. It can serve as a core analytical component in trade monitoring platforms, assisting in policy formulation, international coordination, and food security early-warning systems.

Overall, IVGAE-TAMA-BO offers a scalable and reliable AI-based modeling tool for supporting strategic decision-making in global trade governance.

## 4.4 Limitations and future directions

Despite the promising outcomes achieved in this study, several limitations remain, which also point toward valuable directions for future research.

First, the current framework primarily relies on structural data and a limited set of static economic indicators. It does not yet incorporate complex exogenous variables, such as climatic shocks, transportation costs, or geopolitical instability, which can profoundly influence trade dynamics. Future work may benefit from multi-modal data integration, incorporating sources such as remote sensing imagery, real-time market indices, or textual data from news and policy documents to improve predictive sensitivity under external perturbations. Second, the model currently assumes binary, homogeneous link structures, without explicitly modeling edge directionality, continuous trade volumes, or the presence of multilateral trade agreements. To more accurately capture the multi-layered nature of international trade, subsequent research could extend the IVGAE-TAMA-BO framework to weighted graphs, directed graphs, or even heterogeneous graph neural networks. Third, the temporal modeling strategy in this work is based on fixed-length sliding windows, which may not fully capture seasonal effects or long-term lags that are prevalent in some crop-specific trade patterns. Future studies could explore the incorporation of adaptive temporal attention mechanisms or sequence-aware memory components to provide more flexible and context-aware dynamic modeling. Finally, although the model has been extensively validated on five major crops, its transferability and long-term deployment robustness across a wider range of agricultural commodities and at finer geographic resolutions remains to be fully evaluated. Such efforts will be essential for advancing the real-world applicability of this framework in global food security surveillance and trade resilience assessment.

## 5 Conclusion

In this study, we introduce the dynamic graph neural networks into the context of international food trade, proposed IVGAE-TAMA-BO, a novel dynamic variational graph neural network model for link prediction in global food trade networks. By integrating a variational graph autoencoder with a Trade-Aware Momentum Aggregator (TAMA) and automated Bayesian hyperparameter optimization, the model effectively captures both the temporal evolution and structural persistence of international trade relationships. Extensive experiments on five major crop-specific datasets demonstrate that IVGAE-TAMA consistently outperforms both static and dynamic baselines in terms of AUC and AP metrics, confirming its strong predictive capability and robustness across diverse network structures. Moreover, the incorporation of Bayesian optimization leads to significant performance gains, further enhancing the model's accuracy and generalizability across different trade scenarios. Sensitivity analysis shows that overly long sliding windows tend to degrade accuracy, highlighting the importance of appropriate temporal input length.

Overall, this study provides a robust and scalable modeling framework for link prediction in global food trade networks, offering valuable insights for food security analysis, trade risk assessment, and policy planning.

# Data and Code Availability

The model's code used in this study are publicly available at the Open Science Framework (OSF) repository: https://osf.io/zxek9/.

# Acknowledgements

# Conflicts of Interest

The authors declare no conflicts of interest.